%% file: acl_latex.tex
\pdfoutput=1

\documentclass[11pt]{article}

\usepackage{acl}

\usepackage{times}
\usepackage{latexsym}
\usepackage{booktabs}
\usepackage{multirow}
\usepackage{graphicx}
\usepackage{tabularx}
\usepackage[T1]{fontenc}

\usepackage[utf8]{inputenc}

\usepackage{microtype}

\title{TEAM-Atreides at SemEval-2022 Task 11: On leveraging data augmentation and ensemble to recognize complex Named Entities in Bangla}

\newcommand\blfootnote[1]{%
  \begingroup
  \renewcommand\thefootnote{}\footnote{#1}%
  \addtocounter{footnote}{-1}%
  \endgroup
}

\author{Nazia Tasnim* \and Istiak Shihab* \\
        Department of CSE,
        SUST \\
        \texttt{nimzia.blu@gmail.com}\\
        \texttt{istiak@protonmail.ch}
        \And
  Asif Shahriyar Sushmit \\
  Bengali.Ai \\
  \texttt{sushmit@bengali.ai} \\\AND
  Steven Bethard \\
  School of Information\\
  University of Arizona\\
  Tucson, AZ 85721 \\
  \texttt{bethard@arizona.edu} \\\And
  Farig Sadeque\\
  Department of CSE\\
  BRAC University\\
  Dhaka 1212\\
  \texttt{farig.sadeque@bracu.ac.bd}}

\begin{document}
\maketitle

\input{1_abstract}

\input{2_background}
\input{4_data}

\input{5_methodology}
\input{8_experimental_setup}

\input{6_results}
\input{7_conclusion}


\bibliography{anthology,custom}
\bibliographystyle{acl_natbib}

\end{document}

%% file: 1_abstract.tex
\begin{abstract}
Biological and healthcare domains, artistic works, and organization names can all have nested, overlapping, discontinuous entity mentions that may be syntactically or semantically ambiguous in practice. Traditional sequence tagging algorithms are unable to recognize these complex mentions because they violate the assumptions upon which sequence tagging schemes are founded. In this paper, we describe our contribution to SemEval 2022 Task 11 on identifying such complex named entities. We leveraged an ensemble of ELECTRA-based models exclusively pretrained on the Bangla language with ELECTRA-based monolingual models pretrained on English to achieve competitive performance. Besides providing a system description, we also present the outcomes of our experiments on architectural decisions, dataset augmentations and post-competition findings.
\blfootnote{*These authors contributed equally}
\end{abstract}

%% file: 2_background.tex
\section{Introduction and Related Works}
\label{intro}
The task of identifying and classifying entities in text is known as named entity recognition (NER). Some named entities are easy to distinguish in English since each of their words is capitalized; e.g. "The capital of Bangladesh is Dhaka". In this sentence, both "Bangladesh" and "Dhaka" are capitalized named entities. But there are  other entity mentions that are not simple nouns and are more difficult to recognize.
In the SemEval Task 11:  MultiCoNER Multilingual Complex Named Entity Recognition \cite{multiconer-report}, the organizers concentrated on the more unusual Named Entities, which can be difficult to identify accurately from the text. 

NER tasks have received much attention from the research community due to its crucial role in different NLP problems like information retrieval ~\cite{etzioni2005unsupervised}, Question Answering ~\cite{banko2002askmsr} ~\cite{toral2005improving}, Relation extraction, Entity linking ~\cite{limsopatham2016normalising} and searching ~\cite{pasca2004acquisition}. However, there is such a conceptual difference between an ordinary named entity and a complex named entity that traditional tagging strategies cannot be used to recognize these mentions ~\cite{brown1992class}. Complex NERs can be any language element (single word, abbreviations, imperative clauses, questions) of ambiguous (Multi-type or Overlapping) and non-regular forms (Nested or Discontinuous or Overlapping) ~\cite{ashwini2014targetable}. What makes the task more challenging is, Complex NER is part of the open-domain with ever expanding and emerging entity sets and categories. 

In recent days, Transformer-based models ~\cite{devlin2018bert} ~\cite{liu2019roberta} ~\cite{yang2019xlnet} have been performing as the state-of-the-art ~\cite{yamada2020luke}~\cite{yan2019tener} models in different NER benchmark datasets. Although, Augenstein and colleagues, demonstrate in their paper that these powerful models are only good at picking up the conventional NERs from well formed texts ~\cite{augenstein2017generalisation}, while for complex NERs we still need to integrate external knowledge sources. A recent paper on integrating external sources or Gazetteer features in combination with contextual information, has shown that this can indeed improve performance on complex NER tasks ~\cite{meng2021gemnet}.  Gazetteer-based solutions also show good performance improvements in extracting NERs from both normal and code-mixed webqueries~\cite{fetahu2021gazetteer}.

In tasks like NER, Bangla NLP has not made significant progress. Many linguistic issues arise while training models on Bangla because it is a rich language in terms of both usability and vocabulary ~\cite{ekbal2009named}. In Bangla, there are few markers for tags, such as capitalization ~\cite{karim2019step}. The same words can have a variety of meanings and types of entities. In addition, because Bangla is a somewhat free word order language, words can exist in any place inside a phrase without changing their meaning ~\cite{ekbal2008named}. Affixes that are added to the root word to cause complex inflections can modify the meaning and type of the word as well ~\cite{ekbal2009named}. Despite these issues, transfomer models have been used with considerable success for NER tasks in Bangla ~\cite{bhattacharjee2021banglabert} ~\cite{ashrafi2020banner}.

In this work, we demonstrate our approaches in tackling the concerns raised in the SemEval Task 11, as well as the obstacles posed by the Bangla language's intrinsic complexity. In our proposed architecture, we used a variety of methodologies, primarily focusing on transfer-learning with state-of-the-art deep learning architectures. In particular, we submitted the results obtained from mono-lingual ELECTRA models, while we also ran experiments with non-contextual word embeddings and multilingual language models. 



%% file: 4_data.tex
\section{Dataset Description}
\label{sec: data}
\begin{figure}
     \centering
     \includegraphics[width=\columnwidth]{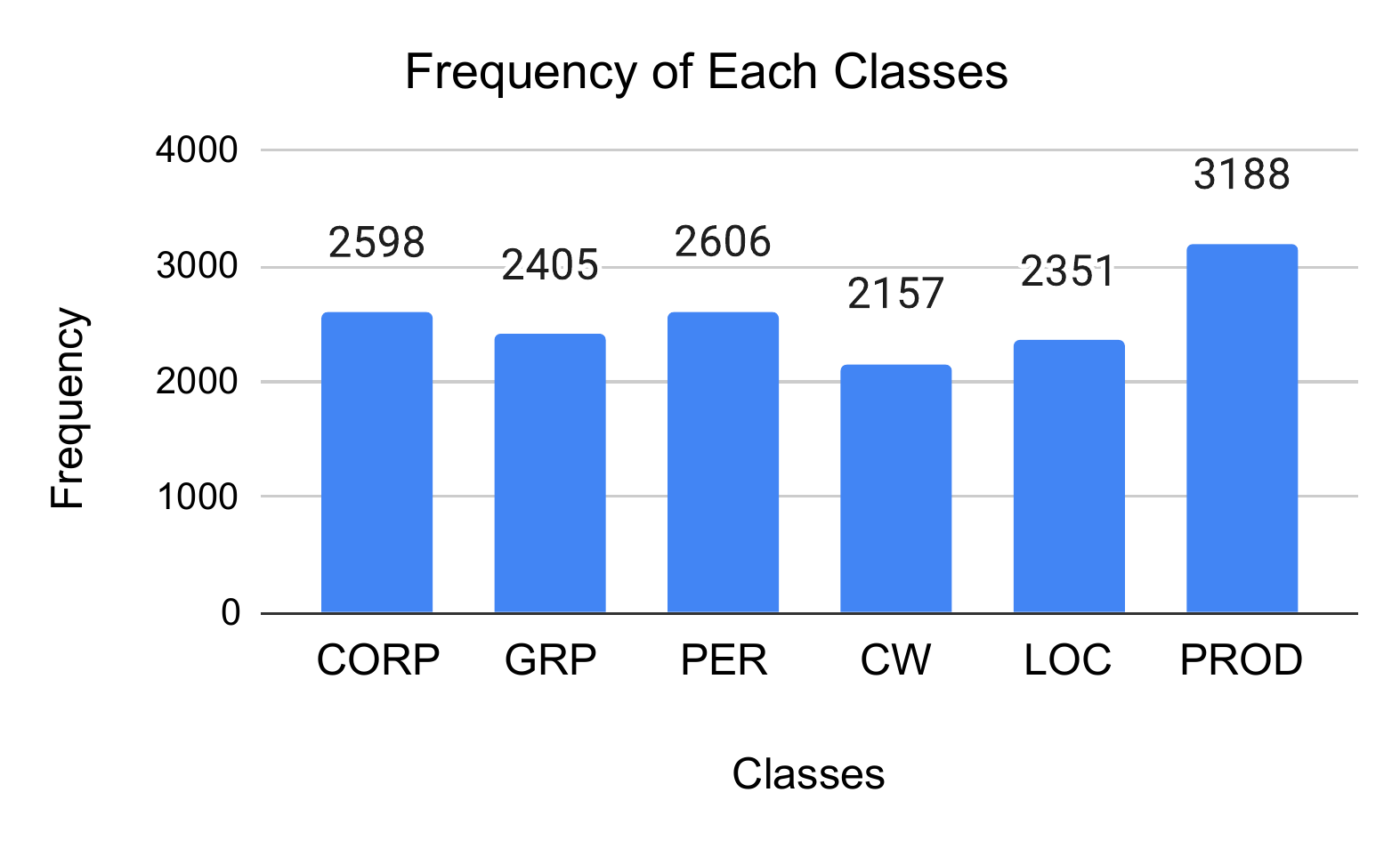}
     \caption{Frequency of each NER Classes}
     \label{fig:freq}
 \end{figure}

 According to the organizers, the data were gathered from Wikipedia and Microsoft Orcas, which included both statements and queries \cite{multiconer-data}. The train set contains about 100 domain adaption instances, whereas the test set has significantly more out-of-domain data to measure out-of-domain performance. The test dataset is a large file of 130k+ sentences, with a preset training dataset of 15300 Bangla sentences and a development dataset of 800 sentences. Other important statistics about the dataset is presented in \ref{tab: dataset}. The distribution of NER classes in the training set is shown in figure \ref{fig:freq}. 
 
\begin{table}[!htp]\centering
\begin{tabular}{lrr}\toprule
\textbf{Type}&\textbf{Frequency}\\\midrule
Train &15300 \\
Dev &800 \\
Test &133119 \\
\midrule
Single Word Tokens &4824 \\
Multi Word Tokens &10481 \\
\bottomrule
\end{tabular}
\caption{Dataset Statistics}
\label{tab: dataset}
\end{table}
 
\begin{table*}[!htp]\centering
\begin{tabular}{lrrrr}\toprule
Language &Dataset Version &Dataset Constituents &Train Set (Sentences) \\\midrule
\multirow{3}{*}{Bangla} &D1 &Bangla &15300 \\
&D2 &Bangla + Hindi(tr.) + Farsi(tr.) &21673 \\
&D3 &Bangla + All(tr.)&82552 \\
\midrule
\multirow{3}{*}{English} &D4 & Bangla(tr.)& 15300\\
 &D5 & Bangla(tr.) + Hindi(tr.) & 30600\\
 &D6 & Bangla(tr.) + Hindi(tr.) + English & 45900\\
\bottomrule
\end{tabular}
\caption{Default and Agumented Dataset Summary}\label{tab: augData}
\end{table*}

 To perform the experiments, we augmented our datasets in several stages. At  first  we token-wise translated a portion of our non-Bangla dataset to Bangla using google translate API\footnote{https://cloud.google.com/translate}. In the first stage, we combined translated Hindi and Farsi dataset with our Bangla dataset, as all three languages come from the Indo-Iranian~\cite{enwiki:1070725920} family. Bangla contains borrowed words from Farsi and it has the same sentence structures as Hindi. In the next step, we combined subsets of translated sentences from all the non-Bangla dataset. This process is repeated for English as well. However, for English, we only combined English, Hindi and Bangla datasets. A summary of our augmented datasets is given in table \ref{tab: augData}.

%% file: 5_methodology.tex





\section{System Description}
The system we proposed for complex Bangla Named Entity Recognition is an ensemble of ELECTRA based models trained on the augmented datasets mentioned in table \ref{tab: augData} and a combination of hyperparameters shown in table \ref{tab: hpms}. The representation of each token is fed into our sequence tagging algorithms, which generate a label for each token. The tag of one token is determined by the attributes of that token in context as well as the tag of the token before it. To execute joint inference, these local decisions are connected together. 

The implementation of our mono-lingual ELECTRA-based systems can broadly be categorized based on the decision of using non-contextual embeddings (word2vec) with a contextual pretrained weight ~\cite{bhattacharjee2021banglabert}. We defined the vanilla token classification system which is largely based on the huggingface token classification scripts \footnote{https://github.com/huggingface/transformers/blob/master/ examples/pytorch/token-classification/run\_ner.py}, as \textit{S1}. The more advanced NER system incorporating non-contextual embedding and optionally, character CNN ~\cite{chiu2016named} and CRF ~\cite{qin2008global} is defined as \textit{S2}. Finally, we developed a majority voting based ensemble scheme, \textit{S3}, to obtain our final prediction for each token. 


\subsection{S1 : Vanilla ELECTRA-based token classification}
\begin{figure}[!h]
    \centering
    \includegraphics[width=\columnwidth]{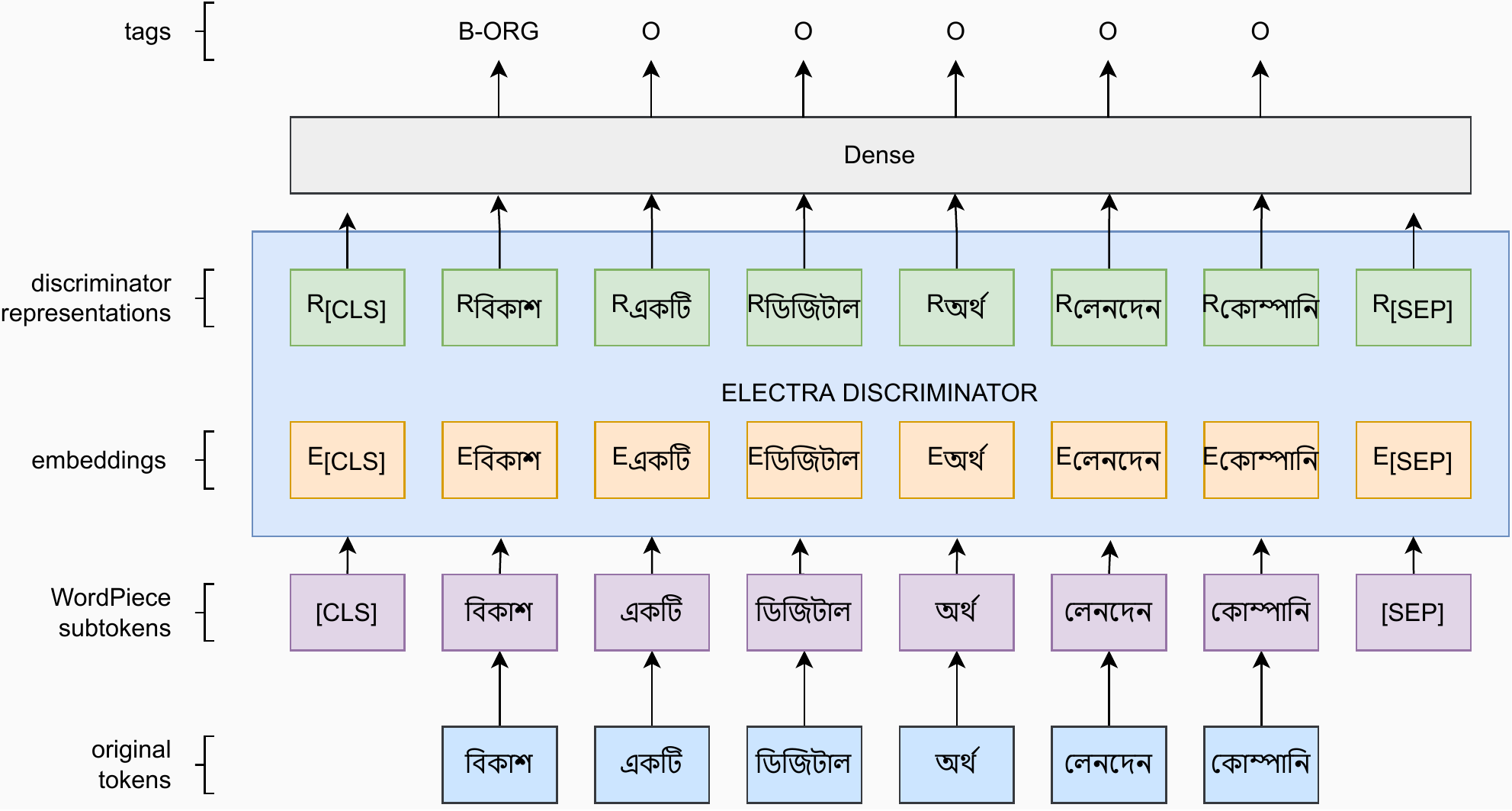}
    \caption{System Overview of \textit{S1}}
    \label{fig:archi_buet}
\end{figure}

The input to \textit{S1} is first normalized using a specific normalization pipeline developed for Bangla mentioned in the ~\cite{hasan2020not} paper. The normalized data is then tokenized and aligned with labels. \textit{S1} has 12 hidden layers, each with 12 attention heads. A standard training loop, with the hyperparameters mentioned in table \ref{tab: hpms} is used in different combinations. Since the original huggingface script does not include an early stopping mechanism, we wrote a custom callback based on evaluation loss and a patience of 5. High-level overview of \textit{S1} is shown in figure ~\ref{fig:archi_buet}.
 
\begin{table}[!htp]\centering
\scriptsize
\begin{tabular}{lrr}\toprule 
System & \multicolumn{2}{c}{Settings}\\
\midrule
\multirow{6}{*}{S1}&\textbf{Tokenizer} &csebuetnlp/banglabert \\
&\textbf{Dropout} &0.1 \\
&\textbf{Batch Size} &[4,8,16] \\
&\textbf{Epoch} &[10,20,30] \\
&\textbf{Patience} &5 \\
&\textbf{Learning Rate} &1.00e-5 \\
&\textbf{Weight Decay} &0.01 \\
\midrule
\multirow{11}{*}{S2}&\textbf{Tokenizer} &csebuetnlp/banglabert \\
&\textbf{Dropout} &[0.0, 0.1, 0.2] \\
&\textbf{LSTM layer} &[2, 4] \\
&\textbf{Batch Size} &[8,16] \\
&\textbf{Epoch} &[30,40,60,100] \\
&\textbf{Patience} &[5,7,10] \\
&\textbf{Use Character CNN} &[True, False] \\
&\textbf{Char CNN Kernel Size} &[3,6,9] \\
&\textbf{Learning Rate} &[1e-05, 5e-o5] \\
&\textbf{Weight Decay } &0.01 \\
&\textbf{Use CRF Layer} &[True, False] \\
\midrule
\multirow{6}{*}{S1.A}&\textbf{Tokenizer} &google/electra-base-discriminator \\
&\textbf{Dropout} &0.1 \\
&\textbf{Batch Size} &64 \\
&\textbf{Epoch} &20 \\
&\textbf{Patience} &5 \\
&\textbf{Learning Rate} &1e-4, 1e-5 \\
\bottomrule
\end{tabular}
\caption{Hyperparameter Settings for \textit{S1 S1.a and S2}}\label{tab: hpms}

\end{table}

\subsubsection{S1.a : Vanilla ELECTRA-based token classification on ENGLISH translated data}
As a preprocessing step for this approach, the input dataset was tokenized and translated to english using Google Translate API. The translated input set is then used with the standard huggingface base Electra model with different combination of hyperparameters, as presented in table \ref{tab: hpms}. We experimented with several token-translated language here with early stopping mechanism at patience of 5. The overall architecture is similar to \textit{S1}.

\subsection{S2: Advanced NER system}
\begin{figure}
    \centering
    \includegraphics[width=\columnwidth]{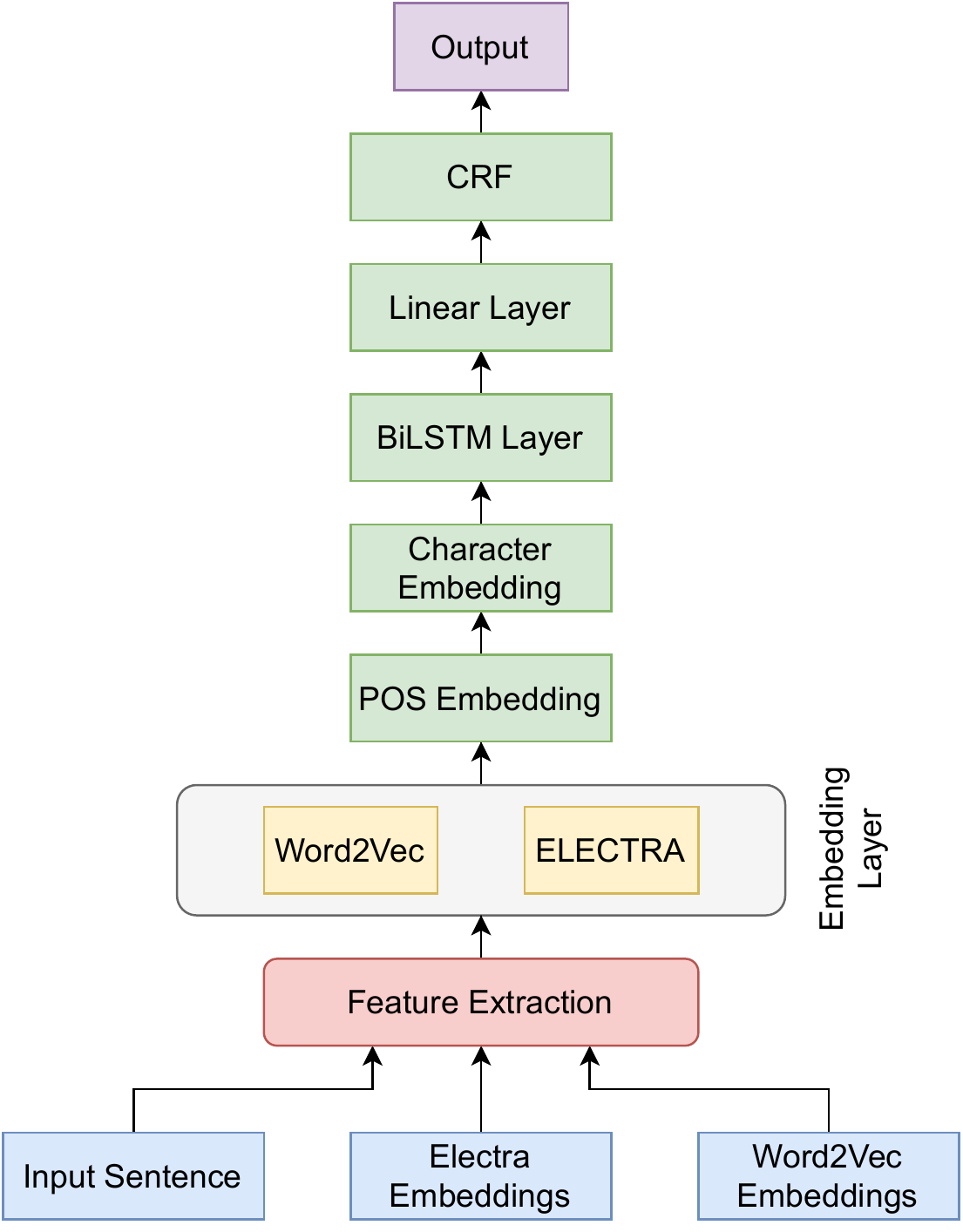}
    \caption{System Overview of \textit{S2}}
    \label{fig:ntagger}
\end{figure}

For this system, character and word level features were first extracted and combined with word2vec and ELECTRA embeddings. To generate the  final embedding these extracted input features passed through a combination of layers including non-contextual embedding layer, contextual pretrained embedding layer, character embedding layer, parts-of-speech (POS) embedding layer, BiLSTM layer and an additional multi-headed attention(MHA) layer. This is projected through a linear layer and optionally goes through a CRF decoding layer to produce the final predictions. This system also included an early stopping mechanism based on evaluation f1 score. An overview of \textit{S2} is presented in figure ~\ref{fig:ntagger}. 

\subsection{S3 : Majority Voting Ensemble}
\begin{figure}
    \centering
    \includegraphics[width=\columnwidth]{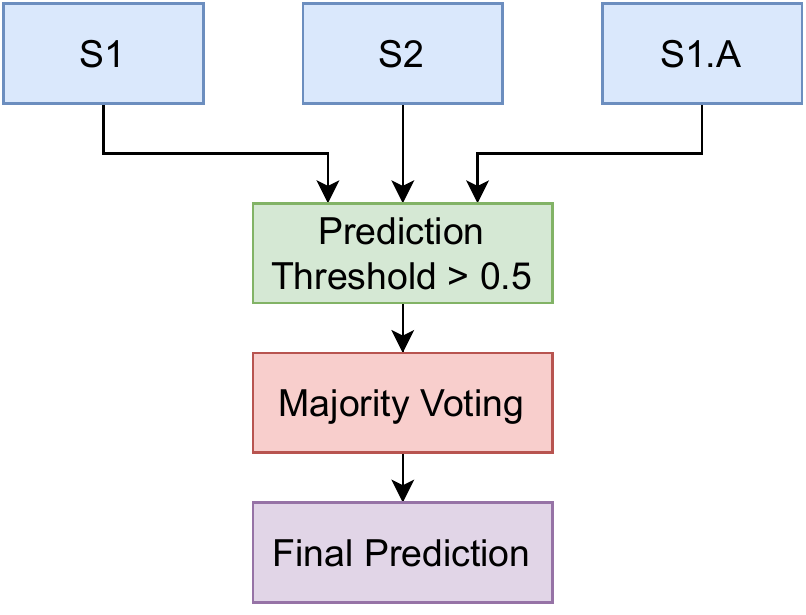}
    \caption{System Overview of \textit{S3}}
    \label{fig:majority}
\end{figure}

The basic concept behind this type of classification is that the final output class is chosen based on the most votes. This ensemble technique has previously been used to overcome the constraints of a single classifier, as presented by the authors in ~\cite{siddiqua2016combining}. 
Before majority voting, we performed a thresholding on the prediction score for each token from each of the 8 models trained using a variety of augmented datasets, pretrained weights, and hyperparameters. We only considered a token label for majority voting if it had a prediction score over 50\%.  Then, we counted the number of times the distilled labels appeared in the set. A label was added to the final list of labels if it appeared in the majority of the models. Overview of the \textit{S3} is shown in ~\ref{fig:majority}.

%% file: 8_experimental_setup.tex





\section{Experimental Setup}
As we have previously discussed in section \ref{sec: data},  we augmented our training data in multiple steps which extended the dataset several times compared to original. We split each version of these dataset into a 70\%-30\% ratio during training. The default dev set containing 800 sentences is used for the final validation, in choosing the best performing model during test phase. We employed accuracy, precision, recall, and F1 score as evaluation metrics, with the macro averaged F1 score as the primary and official metric, as per the benchmark of SemEval 2022 Task 11: MultiCoNER \cite{multiconer-report}. 

We defined each of our best performing model configurations in table \ref{tab: modConfig}. While training both \textit{S1} and \textit{S2} we utilized all versions of the Bangla augmented data. Additionally, to train \textit{S1.a}  we used all versions of the English translated dataset. In table \ref{tab: hpms} we have provided the range of hyperparameters used for each of our systems. The performance of these individual models is also demonstrated in table \ref{tab: indSum}. However, in case of the English models, we have only presented the configuration and prediction score for the best performing model. It should be noted that, these models were submitted for evaluation after competition deadline. 

\begin{table}[!htp]\centering
\begin{tabular}{lrr}\toprule
\textbf{Model Versions} & \\\midrule
\textbf{M1} &S1 + D1 + MHA \\
\textbf{M2} &S1 + D2 \\
\textbf{M3} &S1 + D4 \\
\textbf{M4} &S2 + D1 + CRF \\
\textbf{M5} &S2 + D2 + CRF \\
\textbf{M6} &S2 + D4 + CRF + MHA \\
\textbf{M7} &S2 + D4 + character CNN \\
\textbf{M8} &S1 + D6\\
\bottomrule
\end{tabular}
\caption{Individual Model Configurations}\label{tab: modConfig}
\end{table}

\begin{table}[!htp]\centering
\begin{tabular}{lrrrr}\toprule
\textbf{Attempt} &\textbf{Precision} &\textbf{Recall } &\textbf{macro F1} \\\midrule
M1 &0.4873 &0.3783 &0.402 \\
M2 &0.6121 &0.6053 &0.6072 \\
M3 &0.5437 &0.5135 &0.5248 \\
M4 &0.5184 &0.487 &0.4971 \\
M5 &0.5433 &0.5253 &0.526 \\
M6 &0.5605 &0.5376 &0.5431 \\
M7 &0.5514 &0.5464 &0.5472 \\
M8 &0.5357 &0.5316 &0.5333 \\
\bottomrule
\end{tabular}
\caption{Individual Model Performance Summary}\label{tab: indSum}
\end{table}

%% file: 6_results.tex



\section{Results}
\label{sec: results}
We made 4 submissions during the test phase, by applying majority voting scheme on various combinations of model predictions. The performance of the final ensemble outputs are presented in \ref{tab: ensmble }. As we can observe, the final ensembles of all models performs the highest and it is ranked 8th among all the other teams in Track 11. From the table \ref{tab: leaderboard}, it is visible that our ensemble model does not perform very well in comparison with the top 3 models and in fact, has a difference of over 20\% with the best performing model.

\begin{table}[!htp]\centering
\begin{tabularx}{\columnwidth}{lrrrr}\toprule
\textbf{Models} &\textbf{Precision} &\textbf{Recall } &\textbf{macro F1} \\\midrule
M1 - M3 &0.5924 &0.566 &0.5768 \\
M4 - M7 &0.5926 &0.5449 &0.5597 \\
M1 - M7 &0.5972 &0.5578 &0.5717 \\
All Models &0.6209 &0.5825 &0.5975 \\
\bottomrule
\end{tabularx}
\caption{System Ensemble Summary}\label{tab: ensmble }
\end{table}

\begin{table}[!htp]\centering
\begin{tabular}{lrr}\toprule
\textbf{Team Name} &\textbf{Score} \\\midrule
USTC-NELSLIP (1st) &0.8424 \\
DAMO-NLP(2nd) &0.8351 \\
NetEase.AI (3rd) &0.7088 \\
Sliced (7th) &0.6305 \\
Team Atreides (8th) &0.5975 \\
brotherhood (9th) &0.5863 \\
\bottomrule
\end{tabular}
\caption{Leaderboard for Track-11}\label{tab: leaderboard}

\end{table}

%% file: 7_conclusion.tex
\section{Discussion and Future Directions}
From section \ref{sec: results} we see that, there's hardly any difference among the variations of the \textit{S2} models, while major fluctuations can be observed among the variations of \textit{S1} models. Furthermore, separately grouped ensembles of \textit{S1} and \textit{S2} performs almost identically, with the combined ensemble of \textit{S1} and \textit{S2}. However, the performance improves upon including the predictions from \textit{S1.a} models, which are trained on English translated datasets. Despite this, the final best model is clearly overfitting because it had over 80\% score on the development dataset, 
while performing significantly worse (approximately 60\%) during the test phase of the competition. 
This outcome may be attributed to several factors, including the choice of hyperparameters, dataset augmentations and splitting process, early stopping criteria etc. 
As per the rules of the competition, we only experimented with mono-lingual models to obtain our results. However, we ran the baseline XLM-RoBERTa model which achieves an f1-score of approximately 68\% on the development dataset. 
There are many scopes of expanding this work. For starters, we would like to refine our data augmentation pipeline to generate more well-formed instances. We would explore and compare the performance of cross-lingual and mono-lingual models. We also believe that, the dataset requires further analysis and should receive both quantitative and qualitative error analysis. In addition, we want to do elaborate ablation studies on the components of our systems. In this paper, we have majorly focused on transfer learning and so, in the future, we want to compare the performance of simpler statistical and shallow models with these deep models. Another thing we don't mention empirically in this paper is the class-wise performance of each of our models. From general observation, we find that all the models perform the worst in identifying CW (creative works) tags, while simpler tags like PER (person) and LOC (location) was the easiest to tag. In future, we look forward to investigate more into the reasons behind this behaviors. Finally, we only exploited a simple majority voting based ensemble scheme during this competition. For our future directions, we would also experiment on fusioning the layers of our models to develop a more sophisticated and informed ensembling scheme.

%% file: acl_latex.bbl
\begin{thebibliography}{27}
\expandafter\ifx\csname natexlab\endcsname\relax\def\natexlab#1{#1}\fi

\bibitem[{Ashrafi et~al.(2020)Ashrafi, Mohammad, Mauree, Nijhum, Karim,
  Mohammed, and Momen}]{ashrafi2020banner}
Imranul Ashrafi, Muntasir Mohammad, Arani~Shawkat Mauree, Galib Md~Azraf
  Nijhum, Redwanul Karim, Nabeel Mohammed, and Sifat Momen. 2020.
\newblock Banner: a cost-sensitive contextualized model for bangla named entity
  recognition.
\newblock \emph{IEEE Access}, 8:58206--58226.

\bibitem[{Ashwini and Choi(2014)}]{ashwini2014targetable}
Sandeep Ashwini and Jinho~D Choi. 2014.
\newblock Targetable named entity recognition in social media.
\newblock \emph{arXiv preprint arXiv:1408.0782}.

\bibitem[{Augenstein et~al.(2017)Augenstein, Derczynski, and
  Bontcheva}]{augenstein2017generalisation}
Isabelle Augenstein, Leon Derczynski, and Kalina Bontcheva. 2017.
\newblock Generalisation in named entity recognition: A quantitative analysis.
\newblock \emph{Computer Speech \& Language}, 44:61--83.

\bibitem[{Banko et~al.(2002)Banko, Brill, Dumais, and Lin}]{banko2002askmsr}
Michele Banko, Eric Brill, Susan Dumais, and Jimmy Lin. 2002.
\newblock Askmsr: Question answering using the worldwide web.
\newblock In \emph{Proceedings of 2002 AAAI Spring Symposium on Mining Answers
  from Texts and Knowledge Bases}, pages 7--9.

\bibitem[{Bhattacharjee et~al.(2021)Bhattacharjee, Hasan, Samin, Islam, Rahman,
  Iqbal, and Shahriyar}]{bhattacharjee2021banglabert}
Abhik Bhattacharjee, Tahmid Hasan, Kazi Samin, Md~Saiful Islam, M~Sohel Rahman,
  Anindya Iqbal, and Rifat Shahriyar. 2021.
\newblock Banglabert: Combating embedding barrier in multilingual models for
  low-resource language understanding.
\newblock \emph{arXiv preprint arXiv:2101.00204}.

\bibitem[{Brown et~al.(1992)Brown, Della~Pietra, Desouza, Lai, and
  Mercer}]{brown1992class}
Peter~F Brown, Vincent~J Della~Pietra, Peter~V Desouza, Jennifer~C Lai, and
  Robert~L Mercer. 1992.
\newblock Class-based n-gram models of natural language.
\newblock \emph{Computational linguistics}, 18(4):467--480.

\bibitem[{Chiu and Nichols(2016)}]{chiu2016named}
Jason~PC Chiu and Eric Nichols. 2016.
\newblock Named entity recognition with bidirectional lstm-cnns.
\newblock \emph{Transactions of the association for computational linguistics},
  4:357--370.

\bibitem[{Devlin et~al.(2018)Devlin, Chang, Lee, and
  Toutanova}]{devlin2018bert}
Jacob Devlin, Ming-Wei Chang, Kenton Lee, and Kristina Toutanova. 2018.
\newblock Bert: Pre-training of deep bidirectional transformers for language
  understanding.
\newblock \emph{arXiv preprint arXiv:1810.04805}.

\bibitem[{Ekbal and Bandyopadhyay(2009)}]{ekbal2009named}
Asif Ekbal and Sivaji Bandyopadhyay. 2009.
\newblock Named entity recognition in bengali: A multi-engine approach.
\newblock \emph{Northern European Journal of Language Technology}, 1:26--58.

\bibitem[{Ekbal et~al.(2008)Ekbal, Haque, and Bandyopadhyay}]{ekbal2008named}
Asif Ekbal, Rejwanul Haque, and Sivaji Bandyopadhyay. 2008.
\newblock Named entity recognition in bengali: A conditional random field
  approach.
\newblock In \emph{Proceedings of the Third International Joint Conference on
  Natural Language Processing: Volume-II}.

\bibitem[{Etzioni et~al.(2005)Etzioni, Cafarella, Downey, Popescu, Shaked,
  Soderland, Weld, and Yates}]{etzioni2005unsupervised}
Oren Etzioni, Michael Cafarella, Doug Downey, Ana-Maria Popescu, Tal Shaked,
  Stephen Soderland, Daniel~S Weld, and Alexander Yates. 2005.
\newblock Unsupervised named-entity extraction from the web: An experimental
  study.
\newblock \emph{Artificial intelligence}, 165(1):91--134.

\bibitem[{Fetahu et~al.(2021)Fetahu, Fang, Rokhlenko, and
  Malmasi}]{fetahu2021gazetteer}
Besnik Fetahu, Anjie Fang, Oleg Rokhlenko, and Shervin Malmasi. 2021.
\newblock {Gazetteer Enhanced Named Entity Recognition for Code-Mixed Web
  Queries}.
\newblock In \emph{Proceedings of the 44th International ACM SIGIR Conference
  on Research and Development in Information Retrieval}, pages 1677--1681.

\bibitem[{Hasan et~al.(2020)Hasan, Bhattacharjee, Samin, Hasan, Basak, Rahman,
  and Shahriyar}]{hasan2020not}
Tahmid Hasan, Abhik Bhattacharjee, Kazi Samin, Masum Hasan, Madhusudan Basak,
  M~Sohel Rahman, and Rifat Shahriyar. 2020.
\newblock Not low-resource anymore: Aligner ensembling, batch filtering, and
  new datasets for bengali-english machine translation.
\newblock \emph{arXiv preprint arXiv:2009.09359}.

\bibitem[{Karim et~al.(2019)Karim, Islam, Simanto, Chowdhury, Roy, Al~Neon,
  Hasan, Firoze, Rahman et~al.}]{karim2019step}
Redwanul Karim, MA~Islam, Sazid~Rahman Simanto, Saif~Ahmed Chowdhury, Kalyan
  Roy, Adnan Al~Neon, Md~Hasan, Adnan Firoze, Rashedur~M Rahman, et~al. 2019.
\newblock A step towards information extraction: Named entity recognition in
  bangla using deep learning.
\newblock \emph{Journal of Intelligent \& Fuzzy Systems}, 37(6):7401--7413.

\bibitem[{Limsopatham and Collier(2016)}]{limsopatham2016normalising}
Nut Limsopatham and Nigel Collier. 2016.
\newblock Normalising medical concepts in social media texts by learning
  semantic representation.
\newblock In \emph{Proceedings of the 54th annual meeting of the association
  for computational linguistics (volume 1: long papers)}, pages 1014--1023.

\bibitem[{Liu et~al.(2019)Liu, Ott, Goyal, Du, Joshi, Chen, Levy, Lewis,
  Zettlemoyer, and Stoyanov}]{liu2019roberta}
Yinhan Liu, Myle Ott, Naman Goyal, Jingfei Du, Mandar Joshi, Danqi Chen, Omer
  Levy, Mike Lewis, Luke Zettlemoyer, and Veselin Stoyanov. 2019.
\newblock Roberta: A robustly optimized bert pretraining approach.
\newblock \emph{arXiv preprint arXiv:1907.11692}.

\bibitem[{Malmasi et~al.(2022{\natexlab{a}})Malmasi, Fang, Fetahu, Kar, and
  Rokhlenko}]{multiconer-data}
Shervin Malmasi, Anjie Fang, Besnik Fetahu, Sudipta Kar, and Oleg Rokhlenko.
  2022{\natexlab{a}}.
\newblock {MultiCoNER: a Large-scale Multilingual dataset for Complex Named
  Entity Recognition}.

\bibitem[{Malmasi et~al.(2022{\natexlab{b}})Malmasi, Fang, Fetahu, Kar, and
  Rokhlenko}]{multiconer-report}
Shervin Malmasi, Anjie Fang, Besnik Fetahu, Sudipta Kar, and Oleg Rokhlenko.
  2022{\natexlab{b}}.
\newblock {SemEval-2022 Task 11: Multilingual Complex Named Entity Recognition
  (MultiCoNER)}.
\newblock In \emph{Proceedings of the 16th International Workshop on Semantic
  Evaluation (SemEval-2022)}. Association for Computational Linguistics.

\bibitem[{Meng et~al.(2021)Meng, Fang, Rokhlenko, and Malmasi}]{meng2021gemnet}
Tao Meng, Anjie Fang, Oleg Rokhlenko, and Shervin Malmasi. 2021.
\newblock {GEMNET: Effective gated gazetteer representations for recognizing
  complex entities in low-context input}.
\newblock In \emph{Proceedings of the 2021 Conference of the North American
  Chapter of the Association for Computational Linguistics: Human Language
  Technologies}, pages 1499--1512.

\bibitem[{Pasca(2004)}]{pasca2004acquisition}
Marius Pasca. 2004.
\newblock Acquisition of categorized named entities for web search.
\newblock In \emph{Proceedings of the thirteenth ACM international conference
  on Information and knowledge management}, pages 137--145.

\bibitem[{Qin et~al.(2008)Qin, Liu, Zhang, Wang, and Li}]{qin2008global}
Tao Qin, Tie-Yan Liu, Xu-Dong Zhang, De-Sheng Wang, and Hang Li. 2008.
\newblock Global ranking using continuous conditional random fields.
\newblock \emph{Advances in neural information processing systems}, 21.

\bibitem[{Siddiqua et~al.(2016)Siddiqua, Ahsan, and
  Chy}]{siddiqua2016combining}
Umme~Aymun Siddiqua, Tanveer Ahsan, and Abu~Nowshed Chy. 2016.
\newblock Combining a rule-based classifier with ensemble of feature sets and
  machine learning techniques for sentiment analysis on microblog.
\newblock In \emph{2016 19th international conference on computer and
  information technology (ICCIT)}, pages 304--309. IEEE.

\bibitem[{Toral et~al.(2005)Toral, Noguera, Llopis, and
  Munoz}]{toral2005improving}
Antonio Toral, Elisa Noguera, Fernando Llopis, and Rafael Munoz. 2005.
\newblock Improving question answering using named entity recognition.
\newblock In \emph{International Conference on Application of Natural Language
  to Information Systems}, pages 181--191. Springer.

\bibitem[{{Wikipedia contributors}(2022)}]{enwiki:1070725920}
{Wikipedia contributors}. 2022.
\newblock \href
  {https://en.wikipedia.org/w/index.php?title=Indo-Iranian_languages&oldid=1070725920}
  {Indo-iranian languages --- {Wikipedia}{,} the free encyclopedia}.
\newblock [Online; accessed 28-February-2022].

\bibitem[{Yamada et~al.(2020)Yamada, Asai, Shindo, Takeda, and
  Matsumoto}]{yamada2020luke}
Ikuya Yamada, Akari Asai, Hiroyuki Shindo, Hideaki Takeda, and Yuji Matsumoto.
  2020.
\newblock Luke: deep contextualized entity representations with entity-aware
  self-attention.
\newblock \emph{arXiv preprint arXiv:2010.01057}.

\bibitem[{Yan et~al.(2019)Yan, Deng, Li, and Qiu}]{yan2019tener}
Hang Yan, Bocao Deng, Xiaonan Li, and Xipeng Qiu. 2019.
\newblock Tener: adapting transformer encoder for named entity recognition.
\newblock \emph{arXiv preprint arXiv:1911.04474}.

\bibitem[{Yang et~al.(2019)Yang, Dai, Yang, Carbonell, Salakhutdinov, and
  Le}]{yang2019xlnet}
Zhilin Yang, Zihang Dai, Yiming Yang, Jaime Carbonell, Russ~R Salakhutdinov,
  and Quoc~V Le. 2019.
\newblock Xlnet: Generalized autoregressive pretraining for language
  understanding.
\newblock \emph{Advances in neural information processing systems}, 32.

\end{thebibliography}
